\documentclass[letterpaper]{article} 
\usepackage{aaai23}  
\usepackage{times}  
\usepackage{helvet}  
\usepackage{courier}  
\usepackage[hyphens]{url}  
\usepackage{graphicx} 
\usepackage{natbib}  
\usepackage{caption} 
\usepackage{url}
\usepackage{algorithm}
\usepackage{algorithmic}
\usepackage{cite}
\usepackage{float}
\usepackage{graphicx}				
\nocite{*}								
\usepackage{amssymb}
\usepackage{amsmath}
\usepackage[utf8]{inputenc}
\usepackage[normalem]{ulem}
\usepackage{multirow}
\usepackage[utf8]{inputenc}
\usepackage{tikz}
\usepackage{comment}
\usepackage{pgfplots}
\usepackage{subcaption}
\usepackage{afterpage}
\pgfplotsset{compat=1.17}
\usepackage{newfloat}
\usepackage{listings}
\DeclareCaptionStyle{ruled}{labelfont=normalfont,labelsep=colon,strut=off} 
\DeclareMathOperator*{\argmin}{argmin} 
\floatstyle{ruled}
\newfloat{listing}{tb}{lst}{}
\floatname{listing}{Listing}
\title{GAN-based Domain Inference Attack}
\author{
Yuechun Gu,
Keke Chen
}
\affiliations{
Trustworthy and Intelligent Computing Lab\\
Computer Science Department\\
Marquette University, Milwaukee, WI
}

\begin{document}
\maketitle

\begin{abstract}
Model-based attacks can infer training data information from deep neural network models. These attacks heavily depend on the attacker's knowledge of the application domain, e.g., using it to determine the auxiliary data for model-inversion attacks. However, attackers may not know what the model is used for in practice. We propose a generative adversarial network (GAN) based method to explore likely or similar domains of a target model -- the model domain inference (MDI) attack. For a given target (classification) model, we assume that the attacker knows nothing but the input and output formats and can use the model to derive the prediction for any input in the desired form. Our basic idea is to use the target model to affect a GAN training process for a candidate domain's dataset that is easy to obtain. We find that the target model may distract the training procedure less if the domain is more similar to the target domain. We then measure the distraction level with the distance between GAN-generated datasets, which can be used to rank candidate domains for the target model. Our experiments show that the auxiliary dataset from an MDI top-ranked domain can effectively boost the result of model-inversion attacks. 
\end{abstract}

\section{Introduction}

Numerous companies are applying machine learning in marketing, advertising, and targeting users to improve revenues. Many such machine-learned models depend on sensitive personal or confidential business operational data, raising privacy concerns. So far, most studies about the privacy issue in machine learning focus on problems with training and testing \emph{data privacy} in the stages of model development and inference \cite{chakraborty18}.

In addition to training/testing data privacy, researchers have been wondering how releasing models only can also leak private information in training data. Surprisingly, machine-learned models often remember much more than what we expected \cite{Song17}. Membership inference attacks \cite{Shokri17,Rahman18,Li20,Hui21} can infer which data points were likely used to train the model, and model inversion attacks \cite{Fredrikson15,Wang15,Zhang20,Mahdi21} allow the adversary to approximately reconstruct the training set based on a trained model. 

In all these model-based attacks, the adversaries are assumed to have the prior ``domain knowledge'' of the trained model, which is often defined as the application or task behind the trained model, e.g., a model for facial recognition. The domain knowledge is critical for identifying auxiliary data \cite{Fredrikson15,Zhang20}, e.g., face images, in model inversion attacks. Zhang et al. \cite{Zhang20} show that the attacking accuracy may drop significantly without auxiliary data, and the reconstructed images are not recognizable. Our experiments (Section \ref{Eva}) confirm that this attack-accuracy difference can be up to $\sim30\%$.

Thus, an interesting question is: what if the attacker does not have the desired domain knowledge? In addition to social engineering to explore the domain of a model, does the model itself contain information to infer the domain? Understanding this problem will help us better assess the risk of model-based attacks and design defense mechanisms to protect models.

\textbf{Scope of research and our contributions.}  In this paper, we hold the following minimum assumption about the attacker's prior knowledge. We assume that the target model is a black-box image classification model\footnote{Our current work focuses on the image classification problem. However, it’s possible to extend the study to other learning problems and different types of data, where GAN or other generative methods applies.}. The adversary only knows the input image size and the output probability vector without any knowledge about the domain of the model. However, the adversary can apply the model to any input in the desired format. We assume a bunch of public- or private-domain datasets are available, which can be any image datasets the adversary can find from a set of candidate \emph{landmark domains}. 
        
We focus on a fundamental question which we call \emph{model domain inference}: given a machine learning model, estimate the most similar domain that the model is likely trained on among the available landmark domains. By introducing a method called \emph{latent domain ranking}, we show that even though the model is a black box to the adversaries, it's still possible to derive the target model's relative domain similarities to the landmark domains. Furthermore, we find that the data from top-ranked similar domains work well as the auxiliary data in model-inversion attacks.

A critical problem is estimating the domain similarity between the target model (no target domain data) and any dataset from a candidate domain.  
We develop a generative adversarial network (GAN) based method for this purpose. Specifically, we first train a GAN for a landmark domain with a well-known approach such as Wasserstein GAN (WGAN) \cite{Arjovsky17.1}. After the GAN model converges, we make a snapshot, naming it the \emph{Landmark GAN}. Then, continue to train a copy of the GAN with the latent-domain data and adjust it with the target model via a particular network architecture (Section \ref{GANb}). The target model's response to the GAN-generated images will be fed back to regulate GAN’s incremental training. Finally, we get a \emph{target-model adjusted GAN}.

Intuitively, if the landmark domain is very different from the target domain, the target model's feedback will distract the GAN training and misguide the GAN to generate ``unreal'' records. Otherwise, the target model will be less distractive to the GAN training. The distraction level is measured by the similarity between the two datasets generated by the original landmark GAN and the target-model distracted GAN, respectively. A smaller distance means the landmark GAN is less distracted by the target model, which implies that the landmark domain is more likely related to the target model. Our experimental result shows this method can effectively discover the top domains similar to the target model.

We have also experimented with an alternative method (less effective) for domain inference. We use an existing model inversion method \cite{Fredrikson15} to reconstruct the target domain's training data with a landmark dataset as the auxiliary data. Since the model inversion method depends on the auxiliary data as the hint to the target domain to guide the data reconstruction process, each landmark dataset as the auxiliary data will result in a distinct reconstructed dataset. Then, we evaluate the distance between each pair of the original landmark dataset and its reconstructed. Intuitively, if the landmark domain is similar to the target domain, the auxiliary data should work well to help the model-inversion algorithm generate a good-quality dataset. However, our evaluation shows this method is not as good as the GAN-based method.

One may wonder whether it's possible to mitigate the domain inference attack. Since the attacker implicitly looks at the similarity between the target-domain training data and a landmark dataset (without knowing the target-domain training data), a possible mitigation strategy is to transform the training data. As a result, the attacker tries to find datasets only similar to the transformed training data, which provides an extra layer of protection for the original data. We have identified two recently published data and model disguising methods that meet our needs: InstaHide \cite{Huang20} and RMT \cite{sharma21}. Experimental results show that they can effectively protect models from domain inference attacks with a small sacrifice of model quality.
  
In summary, our contributions include:
\begin{enumerate}
    \item We are the first to study the domain inference problem, which is critical to model-based attacks.    
    \item We have developed a GAN-based domain inference method that can effectively infer the similarity of a dataset to a model. 
    \item We have conducted extensive experiments to validate the proposed method and explored possible mitigation methods to deter the domain inference attack.
\end{enumerate}

We will introduce the notions and definitions in Section \ref{Preliminary}. In Section \ref{Attack}, we will present the threat modeling, the detail of the domain inference attacks, and possible mitigation methods. Then, we show the evaluation result of the attack and mitigation methods in Section \ref{Eva}.

\section{Preliminary}
\label{Preliminary}
We introduce the primary notations, definitions, and necessary background knowledge about GAN and dataset similarity measures in this section.


\subsection{Generative Adversarial Network (GAN)}
A GAN consists of the generator $G$ and the discriminator $D$.

\textbf{Generator.} A generator is a network mapping a random vector to, e.g., a fake image. Its goal is to generate fake data as close to the real data as possible, trying to fool the discriminator. A loss function $L_G=Error(D(G(z)),1)$ serves this purpose.

\textbf{Discriminator.} The discriminator is a binary classifier with a sigmoid function as the activation function. The goal of the discriminator is to correctly separate the generated data labeled as 1 from the real training data labeled as 0. It uses the loss function $L_D=Error(D(x),1)+Error(Dis(G(z)),0)$, where $D()$ is the discriminator output, $x$ is the real data, $z$ is a latent vector, and $G(z)$ is the data generated by generator $G$. The $Error$ function measures the distance between two functional parameters, e.g., cross-entropy or KL-divergence.

\textbf{Optimization.} The overall GAN optimization can also be unified as one function: the generator minimizes, and the discriminator maximizes the loss function $log(D(x))+log(1-D(G(z)))$. The min-max formulation intuitively demonstrates the adversarial process of the competition between the generator and the discriminator. 

\textbf{Wasserstein GAN (WGAN).} The basic GAN suffers from several weaknesses such as slow convergence, vanishing gradient, and model collapse \cite{Arjovsky17.2,Weng19}, which were addressed by the Wasserstein GAN \cite{Arjovsky17.1} later. As a result, WGANs are easier to train and faster to converge, and they also generate better-quality images.

\subsection{Dataset Similarity Measures}
Evaluating the similarity between domains is difficult. The proposed approach will utilize the dataset-level similarity to understand domain similarity. There have been many tools designed for evaluating dataset-level similarity. However, in our approach, the exact dataset similarity is not critical -- instead, we will look at the ranking of similarities to the target domain. We have adopted the Optimal Transport Dataset Distance (OTDD) \cite{Alvarez-Melis20} in this paper and found the result is satisfactory. OTDD is based on a famous distribution transportation problem -- moving one distribution of mass to another as efficiently as possible. It does not depend on pre-trained models, has many good properties, and is relatively easy to compute. 

Fréchet Inception Distance (FID) \cite{Heusel17} is another famous measure popularly used for evaluating the quality of GAN-generated synthetic data. Rather than directly comparing images pixel by pixel between two datasets, FID utilizes the Inception v3 network trained with ImageNet to transform instances from the two datasets. Then the distance is computed based on the mean and standard deviation differences between the two sets of transformed vectors. Since we cannot use the existing Inception-v3 directly for grayscale datasets, we decided not to use FID in this paper. We might explore its use in our future work.
\section{Related Work}
Attacks on training/testing data for machine learning have been extensively discussed in recent few years \cite{chakraborty18}. However, model-based attacks are few, which can be roughly categorized into model inversion attacks and membership inference attacks.

\textbf{Model inversion attacks} try to reconstruct the training examples given access only to a target model and other auxiliary information (e.g., the partial input, the model output, the application domain, and samples from the same domain). Fredrikson et al. \cite{Fredrikson15} demonstrate successful attacks on low-capacity models (e.g., logistic regression and a shallow MLP network) when partial input information was available to the attacker. Hidano et al. \cite{Hidano17} study the scenario without the partial input information. However, their method failed on deep image classifiers. Model inversion attack based on GAN can handle deep neural-network models \cite{Zhang20,Yang19}, which heavily depend on the quality of the auxiliary dataset. However, no study has shown how to identify the auxiliary data when the target model domain is unknown. Our domain inference attack allows the attacker to explore similar domains and identify reliable auxiliary data.

\textbf{Membership inference attacks} try to figure out the likelihood of a record coming from the training data. These attacks assume the targeted record is from the known domain, and some also assume the data distribution is known. Shokri et al. \cite{Shokri17} first propose the concept of membership inference attack, assuming attackers have strong attacking knowledge, which was relaxed by Salem et al. \cite{Salem18}. Long et al. \cite{Long20} propose a method to identify the vulnerable records and models to make the attack more focused. All the above attacks assume attackers know the output probability vector, while Christopher et al. \cite{Choquette-Choo21} explore the label-only membership inference attack, where attackers can only access the output class labels.

\section{Model Domain Inference Attack}
\label{Attack}
As most model inversion attacks depend on known domains, we study the situation when attackers do not know the target model's domain. The core problem here is how to infer likely (or similar) domains for a target model. In this section, we will discuss the threat model, define the domain inference problem, and then present our attacks in detail. We design two attacking methods: the model-inversion-based  method and the GAN-based method.

\subsection{Threat Modeling}

\textbf{Adversarial knowledge.} We hold a minimum assumption that the adversaries can access the target model at least in a black-box manner. However, most model-based attacks depend on the adversarial prior knowledge about the model's domain, which we do not assume the adversaries have. The closest practical setting is the attacker steals the model binary, or breaches the private model inference API, and wants to figure out its secrets. Attackers cannot access the actual training/testing datasets. Otherwise, it’s trivial to infer the problem domain. However, they should know the input image shape and the number of output classes, e.g., 28×28 images and ten output classes. They can choose arbitrary landmark datasets and apply the model to any input data matching the desired format.

\textbf{Attack target and threats.} Domain information is vital for model-based attacks. Knowing the domain allows the adversaries to select an appropriate auxiliary dataset to enhance the attack performance \cite{Zhang20,Yang19}. Most existing model-based attacks assume that domain information is available, which is valid for public model APIs. However, in practice, many models are private or under restricted access. Attackers breach the model access but do not have sufficient domain knowledge about the model. Our study is to explore the limit of what an adversary can do with the reduced adversarial knowledge in these more restricted settings. For instance, can the attacker identify appropriate auxiliary data to enhance model-inversion attacks without knowing the domain? Our experimental results show that the model domain information is probably already embedded in the model itself, and thus model-based attacks may relax the required adversarial knowledge.

\subsection{Definition of Domain Inference}
\label{methods}
The task of domain inference is to estimate the domain of the target model with the help of a bunch of datasets from candidate domains. We define the main concepts as follows.

\textbf{Latent domain.}We define the unknown domain behind the target model as the latent domain. For simplicity, we indicate the latent domain (dataset) as $S_T$.

\textbf{Landmark domains.} We assume the attacker starts with several possible domains $\{S_1, \dots, S_k\}$, which we call landmark domains, to identify the most similar one. Each domain is represented by one dataset -– for simplicity, we reuse the notation, e.g., $S_i$, to represent the dataset in the corresponding domain.

\textbf{Domain Similarity.} Domain similarity is evaluated with the dataset similarity, e.g., the OTDD \cite{Alvarez-Melis20} between the sample datasets from two domains, respectively. As the target model's training data is not accessible, we design two methods to derive approximate domain similarity. The attacker may not need to find out the exact domain similarity -- a good-quality ranking of landmark domains may serve the purpose sufficiently, as we will show.

\subsection{Model-Inversion Based Domain Similarity Estimation}
\begin{figure}[h]
  \centering
    \subcaptionbox{Model-Inversion Reconstruction}[.7\linewidth][c]{%
    \includegraphics[width=\linewidth]{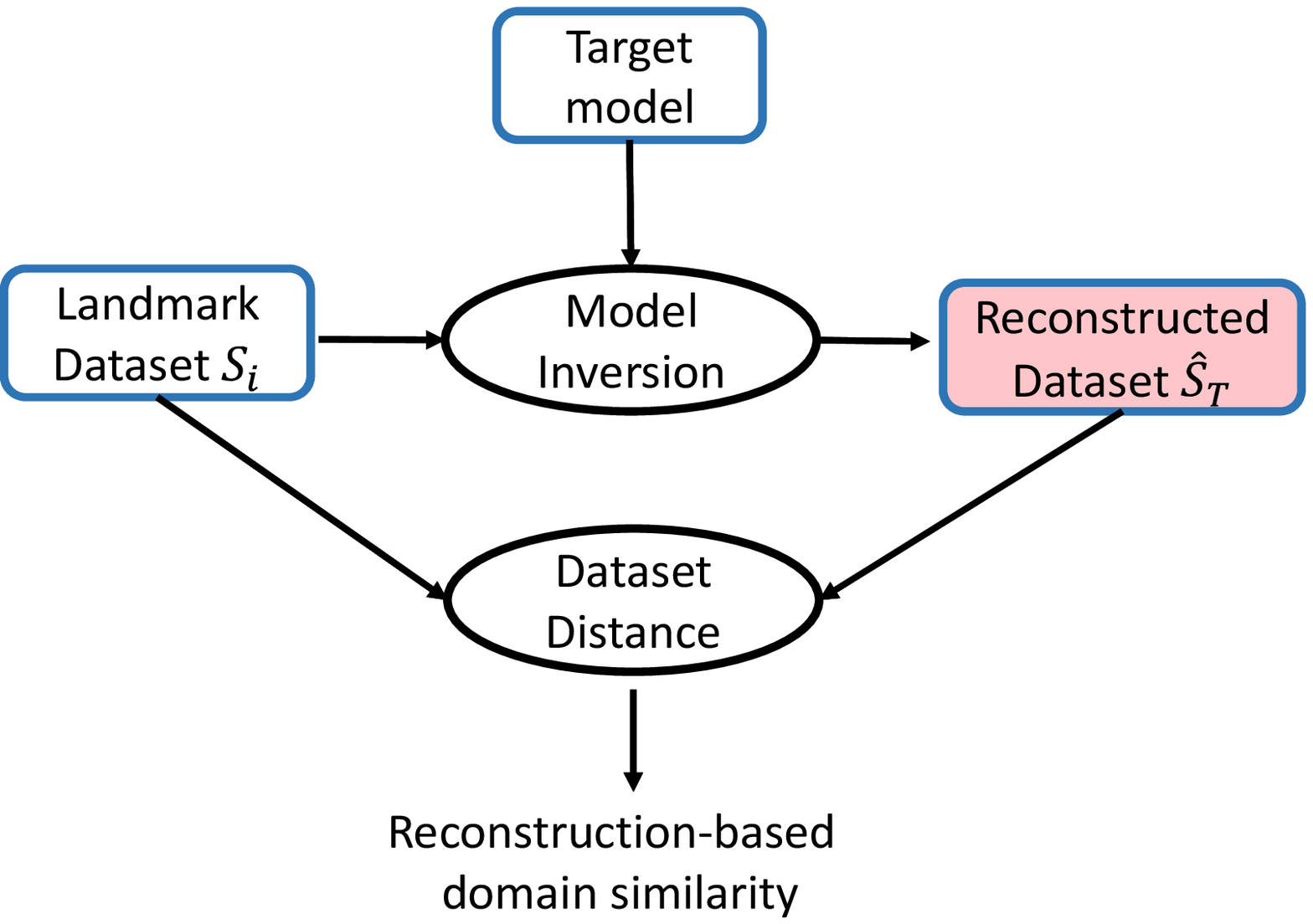}}
  \subcaptionbox{Differential Domain Similarity Estimation}[.7\linewidth][c]{%
    \includegraphics[width=\linewidth]{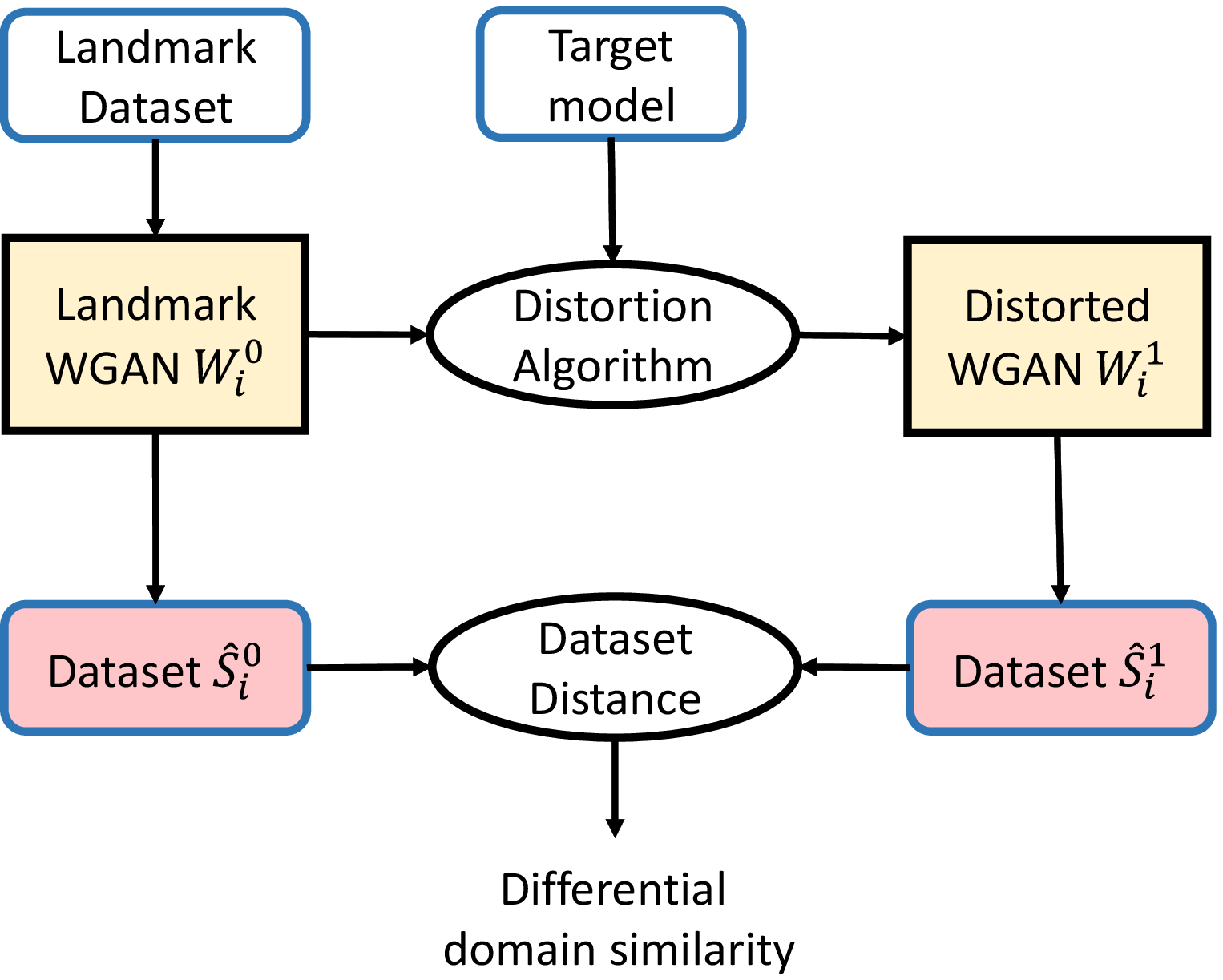}}
   \caption{Architectures of the two domain inference methods}
   \label{F5} 
 \end{figure}

Inspired by the model inversion attack, we design a data-reconstruction-based domain similarity estimation method. The model inversion attack tries to reconstruct the training dataset from the target model with the help of an auxiliary dataset. We can plug in a landmark dataset, e.g., $S_i$, as the auxiliary dataset. The intuition is that if the auxiliary dataset is similar to the target domain data, it will boost the reconstruction process, which in turn generates a dataset, $\hat{S}_{T, i}$, possibly similar to the auxiliary data and the original training data. By measuring the distance between the reconstructed dataset and the landmark dataset, $Dist(\hat{S}_{T,i}, S_i)$, we can infer which landmark dataset is more similar to the latent domain. Figure \ref{F5}(a) shows the steps of this approach. However, in experiments, we find it is not as effective as the GAN-based method we will present next.

\subsection{GAN-based Differential Domain Similarity Estimation}
\label{GANb}
We design a GAN-based method to observe how the target model distracts the landmark domain's GAN training. Intuitively, if the landmark domain is closer to the latent domain than others, the target model will distract the GAN training less. 

As shown in Figure \ref{F5}(b), we first train a \emph{landmark WGAN} $W_i^0$ on each landmark dataset, $S_i$. After training the landmark WGAN, we make a copy of the trained WGAN for the distracted learning process. A specific architecture is used to distrac the continuous training process of the WGAN with the target model, resulting in a \emph{distracted version} $W_i^1$. By observing the distance between the datasets: $\hat{S}_i^0$ generated by $W_i^0$, and  $\hat{S}_i^1$ generated by $W_i^1$, we can derive the \emph{differential domain similarity} as $Dist(\hat{S}_i^0, \hat{S}_i^1)$. We will discuss the important steps as follows.

\begin{figure}[h]
  \centering
    \includegraphics[width=0.7\linewidth]{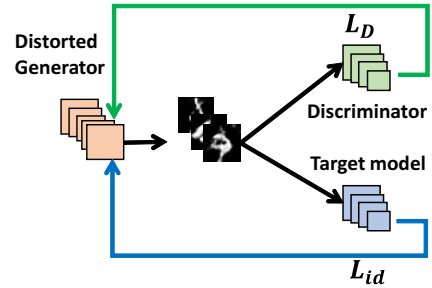}
   \caption{The distracted learning architecture.}
   \label{fig:distortion} 
 \end{figure}

\textbf{Distracted learning architecture.} Starting with a copy of the landmark WGAN, we use the target model to distract the training of the backup WGAN through the ``identity'' loss function $L_{id}$, which describes the probability of the top class in the target model's output: the lower the probability, the more uncertain the prediction is and the more distractions the generator should receive. We build a channel between a trained generator $G$ and the target model $T$, as shown by Figure \ref{fig:distortion}. $G$ is thus influenced by both the identity loss $L_{id}$ and the discriminator's loss $L_{D}$. 

Specifically, for landmark dataset $S_i$ and a record $z$ in the dataset, the target model will accept the generated image $G(z)$ and feedback an identity loss defined as $L_{id}(z) = - log[T(G(z))]$ where $T(G(z))$ is the probability of the predicted top class by the target model $T$. Meanwhile, the discriminator feeds back a prior loss to $G$: $L_{D}=-D(G(z))$. Finally, we solve the following optimization for the generator G: 
\[\hat{z} = \argmin_z L_{D}(z)+\lambda L_{id}(z), 
\]
where $\lambda$ is a predefined parameter representing the influence of the target model.

Intuitively, the loss $L_D$ penalizes unrealistic images, while the identity loss encourages the generated images to have higher prediction top-class likelihood under the targeted model. This distracted learning process tries to guide the generated image $\hat{z}$ towards the latent dataset $S_T$. The image will be distorted more if $S_i$ is less similar to $S_T$ and change less otherwise.

\textbf{Distance measurement.} We have used the datasets generated by the two WGANs, respectively, for distance evaluation. One may wonder why not use the original landmark dataset $S_i$, i.e., computing $Dist(S_i, \hat{S}_i^1)$ instead. The problem is that the WGAN training performs differently for different datasets. It works less effectively for some datasets, e.g., CIFAR10, than others, such as MNIST. To eliminate the quality difference caused by the underlying WGAN training, we use the distance between $\hat{S}_i^0$ and $\hat{S}_i^1$ instead, which more accurately measures the differential effect caused by the target model.

\subsection{Potential Mitigation Methods}
\label{miti}
We look at various ways to protect target models and datasets and mitigate the domain inference attack. 

The first approach is to prevent adversarial access to models and data. CryptoNets \cite{gilad16} uses homomorphic encryption for encrypting data and models in model inference to avoid leaking models and testing examples to adversaries. Several cryptographic protocols have implemented confidential model training, such as SecureML \cite{Mohassel17} and ConfidentialBoost \cite{Sharma19SecureBoost}. They typically use hybrid constructions of homomorphic encryption and secure multiparty computation primitives. More recently, trusted execution environments (TEEs), such as Intel SGX \cite{sgx-explained}, have been used in confidential machine learning. However, TEEs have not been implemented in GPUs yet, and thus the performance of TEE-based deep learning is not satisfactory \cite{Tramer19}.

We also noticed the data and model disguising methods, including InstaHide \cite{Huang20} and DisguisedNets \cite{sharma21}. Both methods require transforming the training data, which results in models that work only on the transformed data. They share a unique benefit compared to other methods: existing GPU-accelerated model training methods can be applied to the transformed data without modification.

InstaHide provides a randomized training data transformation method, which mixes up each private training image with randomly selected and weighted private and public ones. The sign of each mixed-up pixel is also randomly flipped. The transformed images can be directly used to train models. The same transformation method is applied when the model is used for inference.

A simple method of DisguisedNets, randomized multiplicative transformation (RMT), uses a different approach. It partitions each training image into multiple blocks with a pre-defined scheme. A randomly generated invertible transformation matrix corresponds to a block position and is shared by all images serving as a secret key. It then transforms the block at each position with the corresponding secret matrix, e.g., using a noise-added linear transformation. For example, we can partition a 32x32 image into 64 4x4 blocks. Each of the 64 positions uses a 4x4 randomly generated matrix as the key. The block-wise transformation is applied to each image.

\pgfplotstableread{
0   0.882   0.013       
1   0.878   0.009       
2   0.879   0.014       
3   0.782   0.007       
4   0.693   0.009       
5   0.721   0.011       
6   0.627   0.005       
7   0.605   0.007       
}\datasetcor

\pgfplotstableread{
0   0.124   0.008   0.009   0.314   0.012   0.013 0.242 0.024 0.427 0.013
1   0.112   0.009   0.004   0.289   0.007   0.015 0.215 0.021 0.393 0.011
2   0.117   0.004   0.003   0.277   0.011   0.011 0.198 0.014 0.386 0.019
3   0.112   0.005   0.005   0.231   0.013   0.012 0.183 0.015 0.299 0.008
4   0.092   0.007   0.006   0.237   0.007   0.008 0.156 0.018 0.314 0.009
5   0.083   0.005   0.007   0.178   0.009   0.005 0.168 0.012 0.227 0.016
6   0.085   0.003   0.009   0.087   0.011   0.014 0.085 0.003 0.143 0.013
7   0.079   0.002   0.007   0.081   0.014   0.009 0.077 0.002 0.135 0.017
}\datasetf

\begin{figure}[t]
\centering
\resizebox{0.7\linewidth}{!}{
\begin{tikzpicture}[font=\small]
\begin{axis}[ybar,
        width=\linewidth,
        bar width=4pt,
        ymin=0,
        ymax=0.7,        
        xtick=data,
        label style={font=\small},
        tick label style={font=\small}  ,
        xticklabels = {M,E,FM,C,L,Em,C10,C100},
        major x tick style = {opacity=0},
        minor x tick num = 1,
        minor tick length=2ex,
        legend pos=north east, style={font=\small}
        ]
        \addplot[draw=black,fill=black!0,error bars/.cd,y dir=both,y explicit] 
    table[x index=0,y index=1,y error plus index=2,y error minus index=2] \datasetf; 
        \addplot[draw=black,fill=black!20,error bars/.cd,y dir=both,y explicit] 
    table[x index=0,y index=7,y error plus index=8,y error minus index=8] \datasetf; 
\addplot[draw=black,fill=black!40,error bars/.cd,y dir=both,y explicit] 
    table[x index=0,y index=4,y error plus index=5,y error minus index=5] \datasetf; 
    \addplot[draw=black,fill=black!60,error bars/.cd,y dir=both,y explicit] 
    table[x index=0,y index=9,y error plus index=10,y error minus index=10] \datasetf; 
    
    \legend{Without Aux, Rank 3 Dataset as Aux, Rank 2 Dataset as Aux, Rank1 Dataset as Aux}
\end{axis}
\end{tikzpicture}}
\caption{The effect of top-ranked datasets as the auxiliary data for the model-inversion attack. The better ranked the dataset, the more contribution it makes to the attack. Datasets: M, E, FM, C, L, Em, C10, C100 represent MNIST, EMNIST, Fashion-MNIST, Clothing, Emotion, CIFAR10, CIFAR100, respectively.} 
\label{F8}
\end{figure}
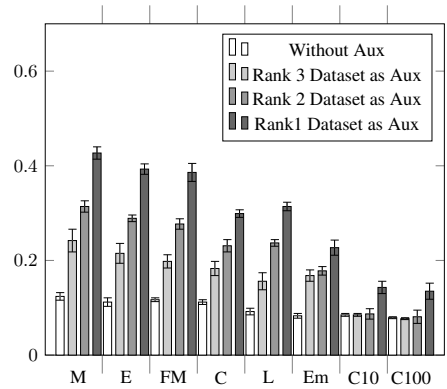

Both InstaHide and RMT can preserve model accuracy relatively well –-  about 2\% - 7\% reduction with a simple network like ResNet-18 in our experiments. More sophisticated networks can preserve model quality better. As the models are trained on the transformed datasets, the domain inference attack and even model-inversion attacks do not work anymore. We will show whether they can protect models from domain inference attacks in experiments. 

While models trained with InstaHide or RMT perturbed data are safe from the MDI attack, users should be aware that InstaHide and RMT methods are subject to other attacks. In particular, an attack using image clustering and pixel sign removing \cite{Carlini21} works effectively on InstaHide. The regression attack can also be a concern on RMT \cite{sharma21} if the attacker collects a sufficient number of original-transformed image pairs, which, however, contradict the MDI's threat model. Readers may adopt these methods cautiously depending on the threat model acceptable to a specific application.

\section{Experiments}
\label{Eva}
In this section, we conduct experiments to show that: (1) How auxiliary data enhances the model inversion attack; (2)  How effective our MDI methods are; and (3) How the suggested mitigation methods work against the MDI attack.
\pgfplotstableread{
0   0.834 0.053 0.024 0.917 0.031 0.003
1   0.817 0.062 0.003 0.892 0.024 0.004
2   0.767 0.066 0.015 0.875 0.032 0.007
3   0.776 0.051 0.011 0.872 0.015 0.005
4   0.814 0.035 0.006 0.863 0.035 0.003
5   0.791 0.042 0.007  0.880 0.029  0.006
6   0.623 0.063 0.023 0.871 0.041 0.021
7   0.617 0.066 0.025 0.782 0.034   0.004  
}\datasetna

\pgfplotstableread{
0   0.679   0.056   0.021   0.914   0.029   0.009
1   0.623   0.059   0.015   0.894   0.025   0.012
2   0.628   0.072   0.008   0.918   0.027   0.003
3   0.599   0.062   0.011   0.907   0.018   0.01
4   0.538   0.061   0.011   0.879   0.022   0.008
5   0.573   0.053   0.013   0.923   0.029   0.002
6   0.526   0.041   0.021   0.804   0.038   0.012
7   0.515   0.053   0.036   0.767   0.034   0.008
}\datasetnb

\pgfplotstableread{
0   0.647   0.049   0.019   0.897   0.022   0.013
1   0.658   0.052   0.011   0.902   0.028   0.007
2   0.617   0.042   0.009   0.884   0.026   0.003
3   0.622   0.054   0.013   0.872   0.022   0.003
4   0.582   0.062   0.017   0.857   0.037   0.005
5   0.592   0.042   0.021   0.918   0.014   0.009
6   0.539   0.067   0.017   0.857   0.042   0.012
7   0.523   0.059   0.028   0.821   0.047   0.003
}\datasetnc

\begin{figure*}[t]
\centering
\resizebox{0.3\linewidth}{!}{
\begin{tikzpicture}[font=\small]
\begin{axis}[ybar,
        width=0.5\textwidth,
        bar width=4pt,
        ymin=0.4,
        ymax=1,           
        ylabel={NDCG@1},
        xlabel={Datasets},
        xtick=data,
        label style={font=\small},
        tick label style={font=\small}  ,
        xticklabels = {M,E,FM,C,L,Em,C10,C100},
        major x tick style = {opacity=0},
        minor x tick num = 1,
        minor tick length=2ex,
        legend pos=south west
        ]

\addplot[draw=black,fill=black!0,error bars/.cd,y dir=both,y explicit] 
    table[x index=0,y index=1,y error plus index=2,y error minus index=2] \datasetna; 
\addplot[draw=black,fill=black!40,error bars/.cd,y dir=both,y explicit] 
    table[x index=0,y index=4,y error plus index=5,y error minus index=5] \datasetna; 
  \legend{Reconstruction, GAN-based}
\end{axis}
\end{tikzpicture}}
\centering
\resizebox{0.3\linewidth}{!}{
\begin{tikzpicture}
\begin{axis}[ybar,
        width=0.5\textwidth,
        bar width=4pt,
        ymin=0.4,
        ymax=1,        
        ylabel={NDCG@2},
        xlabel={Datasets},
        xtick=data,
        label style={font=\small},
        tick label style={font=\small}  ,
        xticklabels = {M,E,FM,C,L,Em,C10,C100},
        major x tick style = {opacity=0},
        minor x tick num = 1,
        minor tick length=2ex,
        legend pos=south east
        ]

\addplot[draw=black,fill=black!0,error bars/.cd,y dir=both,y explicit] 
    table[x index=0,y index=1,y error plus index=2,y error minus index=2] \datasetnb; 
\addplot[draw=black,fill=black!40,error bars/.cd,y dir=both,y explicit] 
    table[x index=0,y index=4,y error plus index=5,y error minus index=5] \datasetnb; 
\end{axis}
\end{tikzpicture}}  
\resizebox{0.3\linewidth}{!}{
\begin{tikzpicture}
\begin{axis}[ybar,
        width=0.5\textwidth,
        bar width=4pt,
        ymin=0.4,
        ymax=1,        
        ylabel={NDCG@3},
        xlabel={Datasets},
        xtick=data,
        label style={font=\small},
        tick label style={font=\small}  ,
        xticklabels = {M,E,FM,C,L,Em,C10,C100},
        major x tick style = {opacity=0},
        minor x tick num = 1,
        minor tick length=2ex,
        legend pos=south west
        ]
\addplot[draw=black,fill=black!0,error bars/.cd,y dir=both,y explicit] 
    table[x index=0,y index=1,y error plus index=2,y error minus index=2] \datasetnc; 
\addplot[draw=black,fill=black!40,error bars/.cd,y dir=both,y explicit] 
    table[x index=0,y index=4,y error plus index=5,y error minus index=5] \datasetnc; 

\end{axis}
\end{tikzpicture}}
\caption{The GAN-based method performs consistently better than the reconstruction method.}
\label{F9}
\end{figure*}
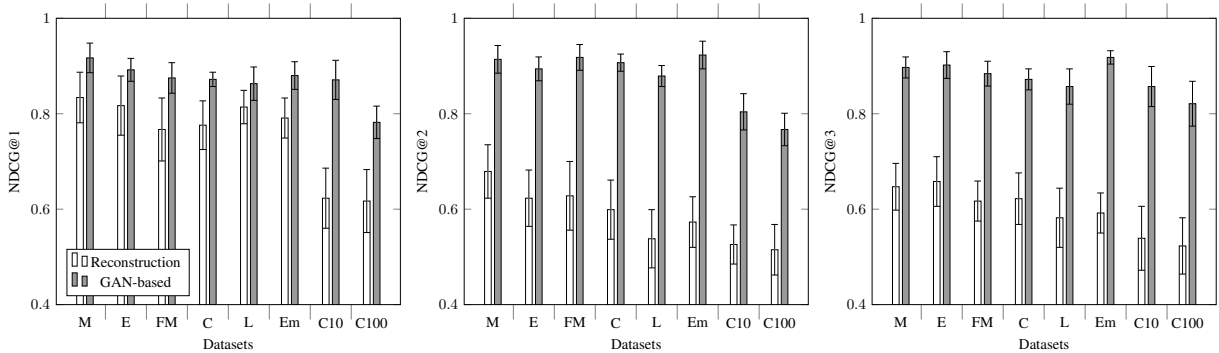

\pgfplotstableread{
0   0.712 0.044 0.024 0.917 0.031 0.003
1   0.676 0.021 0.003 0.892 0.024 0.004
2   0.733 0.052 0.015 0.875 0.032 0.007
3   0.644 0.027 0.011 0.872 0.015 0.005
4   0.636 0.023 0.006 0.863 0.035 0.003
5   0.689 0.013 0.007  0.880 0.029  0.006
6   0.572 0.053 0.023 0.871 0.041 0.021
7   0.555 0.035 0.025 0.782 0.034   0.004  
}\dataseta

\pgfplotstableread{
0   0.677   0.041   0.021   0.914   0.029   0.009
1   0.599   0.025   0.015   0.894   0.025   0.012
2   0.623   0.038   0.008   0.918   0.027   0.003
3   0.523   0.041   0.011   0.907   0.018   0.01
4   0.481   0.031   0.011   0.879   0.022   0.008
5   0.578   0.023   0.013   0.923   0.029   0.002
6   0.592   0.041   0.021   0.804   0.038   0.012
7   0.502   0.033   0.036   0.767   0.034   0.008
}\datasetb

\pgfplotstableread{
0   0.682   0.042   0.019   0.897   0.022   0.013
1   0.644   0.037   0.011   0.902   0.028   0.007
2   0.728   0.047   0.009   0.884   0.026   0.003
3   0.563   0.038   0.013   0.872   0.022   0.003
4   0.621   0.077   0.017   0.857   0.037   0.005
5   0.667   0.039   0.021   0.918   0.014   0.009
6   0.589   0.034   0.017   0.857   0.042   0.012
7   0.553   0.048   0.028   0.821   0.047   0.003
}\datasetc

\pgfplotstableread{
0   0.517 0.231 0.493 0.193  0.917 0.031 
1   0.489 0.276 0.514 0.136 0.892 0.024 
2   0.521 0.283 0.519 0.150 0.875 0.032 
3   0.477 0.191 0.524 0.188 0.872 0.015 
4   0.453 0.142 0.504 0.203 0.863 0.035 
5   0.483 0.171 0.473 0.221 0.880 0.029  
6   0.511 0.237 0.521 0.216 0.871 0.041
7   0.492 0.186 0.447 0.195 0.782 0.034 
}\datasets

\pgfplotstableread{
0   102.6   14.7    14.7
1   97.5    15.8    15.8
2   138.9   10.2    10.2
3   112.3   10.7    10.7
4   92.6    14.2    14.2
5   123.12  12.37   12.37
6   227.39  21.3    21.3
7   203.42  17.2    17.2
}\datasete

\pgfplotstableread{
0   0.954 0.014 0.909  0.005 0.889  0.004
1   0.947 0.012  0.899 0.004 0.902  0.003
2   0.933 0.008  0.871 0.005 0.879 0.005
3   0.907 0.011  0.847 0.003 0.851 0.007
4   0.916 0.008  0.855 0.005 0.843 0.003
5   0.898 0.007  0.852 0.002 0.856  0.005
6   0.903 0.011  0.842 0.006 0.849 0.005
7   0.747 0.009  0.699 0.011 0.682 0.007
}\datasetdown

\begin{figure*}[t]
\centering
\resizebox{0.3\linewidth}{!}{
\begin{tikzpicture}[font=\small]
\begin{axis}[ybar,
        width=0.5\textwidth,
        bar width=4pt,
        ymin=0.4,
        ymax=1,           
        ylabel={NDCG@1},
        xlabel={Datasets},
        xtick=data,
        label style={font=\small},
        tick label style={font=\small}  ,
        xticklabels = {M,E,FM,C,L,Em,C10,C100},
        major x tick style = {opacity=0},
        minor x tick num = 1,
        minor tick length=2ex,
        legend pos=south west
        ]

\addplot[draw=black,fill=black!0,error bars/.cd,y dir=both,y explicit] 
    table[x index=0,y index=1,y error plus index=2,y error minus index=2] \dataseta; 
\addplot[draw=black,fill=black!40,error bars/.cd,y dir=both,y explicit] 
    table[x index=0,y index=4,y error plus index=5,y error minus index=5] \dataseta; 
    \legend{Alternative similarity, Differential similarity}
\end{axis}
\end{tikzpicture}}
\centering
\resizebox{0.3\linewidth}{!}{
\begin{tikzpicture}
\begin{axis}[ybar,
        width=0.5\textwidth,
        bar width=4pt,
        ymin=0.4,
        ymax=1,        
        ylabel={NDCG@2},
        xlabel={Datasets},
        xtick=data,
        label style={font=\small},
        tick label style={font=\small}  ,
        xticklabels = {M,E,FM,C,L,Em,C10,C100},
        major x tick style = {opacity=0},
        minor x tick num = 1,
        minor tick length=2ex,
        legend pos=south east
        ]

\addplot[draw=black,fill=black!0,error bars/.cd,y dir=both,y explicit] 
    table[x index=0,y index=1,y error plus index=2,y error minus index=2] \datasetb; 
\addplot[draw=black,fill=black!40,error bars/.cd,y dir=both,y explicit] 
    table[x index=0,y index=4,y error plus index=5,y error minus index=5] \datasetb; 
\end{axis}
\end{tikzpicture}}  
\resizebox{0.3\linewidth}{!}{
\begin{tikzpicture}
\begin{axis}[ybar,
        width=0.5\textwidth,
        bar width=4pt,
        ymin=0.4,
        ymax=1,        
        ylabel={NDCG@3},
        xlabel={Datasets},
        xtick=data,
        label style={font=\small},
        tick label style={font=\small}  ,
        xticklabels = {M,E,FM,C,L,Em,C10,C100},
        major x tick style = {opacity=0},
        minor x tick num = 1,
        minor tick length=2ex,
        legend pos=south west
        ]
\addplot[draw=black,fill=black!0,error bars/.cd,y dir=both,y explicit] 
    table[x index=0,y index=1,y error plus index=2,y error minus index=2] \datasetc; 
\addplot[draw=black,fill=black!40,error bars/.cd,y dir=both,y explicit] 
    table[x index=0,y index=4,y error plus index=5,y error minus index=5] \datasetc; 
\end{axis}
\end{tikzpicture}}
\caption{Differential similarity is significantly better than the alternative similarity measure for the GAN-based method.}
\label{F2}
\end{figure*}
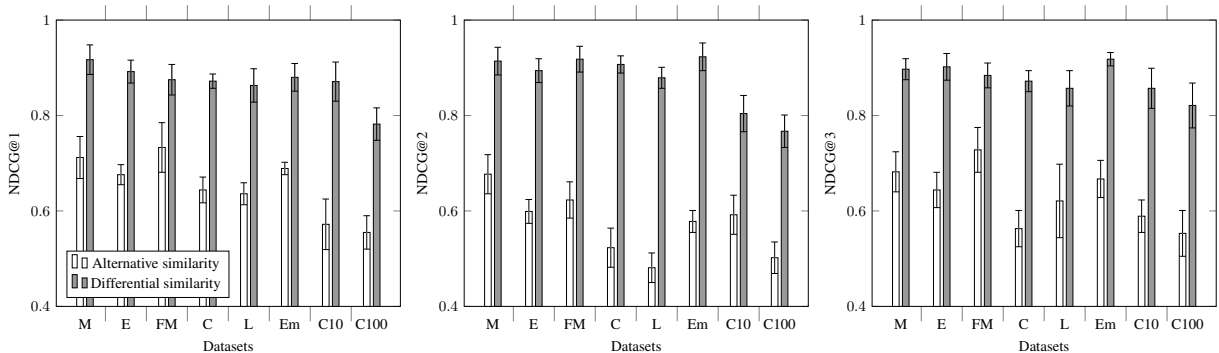

\begin{figure*}[h]
\centering
\resizebox{0.3\linewidth}{!}{
\begin{tikzpicture}[font=\small]
\begin{axis}[ybar,
        width=0.6\textwidth,
        bar width=8pt,
        ymin=50,
        ymax=250,        
        ylabel={$Distance^{WGAN}$},
        xlabel={Datasets},
        xtick=data,
        label style={font=\small},
        tick label style={font=\small}  ,
        xticklabels = {M,E,FM,C,L,Em,C10,C100},
        major x tick style = {opacity=0},
        minor x tick num = 1,
        minor tick length=2ex,
        legend pos=north east
        ]
\addplot[draw=black,fill=black!40,error bars/.cd,y dir=both,y explicit] 
    table[x index=0,y index=1,y error plus index=2,y error minus index=3] \datasete; 
\end{axis}
\end{tikzpicture}
}
\resizebox{0.3\linewidth}{!}{
\begin{tikzpicture}[font=\small]
\begin{axis}[ybar,
        width=0.6\textwidth,
        bar width=4pt,
        ymin=0,
        ymax=1.0,        
        ylabel={NDCG@1},
        xlabel={Datasets},
        xtick=data,
        label style={font=\small},
        tick label style={font=\small}  ,
        xticklabels = {M,E,FM,C,L,Em,C10,C100},
        major x tick style = {opacity=0},
        minor x tick num = 1,
        minor tick length=2ex,
        legend pos=south west 
        ]

\addplot[draw=black,fill=black!0,error bars/.cd,y dir=both,y explicit] 
    table[x index=0,y index=1,y error plus index=2,y error minus index=2] \datasets; 
   \addplot[draw=black,fill=black!20,error bars/.cd,y dir=both,y explicit] 
    table[x index=0,y index=3,y error plus index=4,y error minus index=4] \datasets; 
\addplot[draw=black,fill=black!40,error bars/.cd,y dir=both,y explicit] 
    table[x index=0,y index=5,y error plus index=6,y error minus index=6] \datasets; 
    \legend{RMT, InstaHide, No mitigation}
\end{axis}
\end{tikzpicture}}
\resizebox{0.3\linewidth}{!}{
\begin{tikzpicture}[font=\small]
\begin{axis}[ybar,
        width=0.6\textwidth,
        bar width=4pt,
        ymin=0.6,
        ymax=1,        
        ylabel={Model Quality},
        xtick=data,
        xlabel={Datasets},
        label style={font=\small},
        tick label style={font=\small}  ,
        xticklabels = {M,E,FM,C,L,Em,C10,C100},
        major x tick style = {opacity=0},
        minor x tick num = 1,
        minor tick length=2ex,
        legend pos= south west
        ]
   \addplot[draw=black,fill=black!0,error bars/.cd,y dir=both,y explicit] 
    table[x index=0,y index=3,y error plus index=4,y error minus index=4] \datasetdown; 
\addplot[draw=black,fill=black!20,error bars/.cd,y dir=both,y explicit] 
    table[x index=0,y index=5,y error plus index=6,y error minus index=6] \datasetdown; 
    \addplot[draw=black,fill=black!40,error bars/.cd,y dir=both,y explicit] 
    table[x index=0,y index=1,y error plus index=2,y error minus index=2] \datasetdown; 
    \legend{RMT, InstaHide, No mitigation}
\end{axis}
\end{tikzpicture}}
\caption{(left) $Distance^{WGAN}$ means the distance between the WGAN generated data and the original data. The larger the distance, the worse quality. CIFAR datasets have much lower quality WGANs than others. (mid) InstaHide and RMT can significantly improve the resilience to the MDI attack. (right) However, InstaHide and RMT also slightly reduce model quality, implying a trade-off between attack resilience and model quality.} 
  \label{fig:wgan-quality-and-mitigation}
\end{figure*}
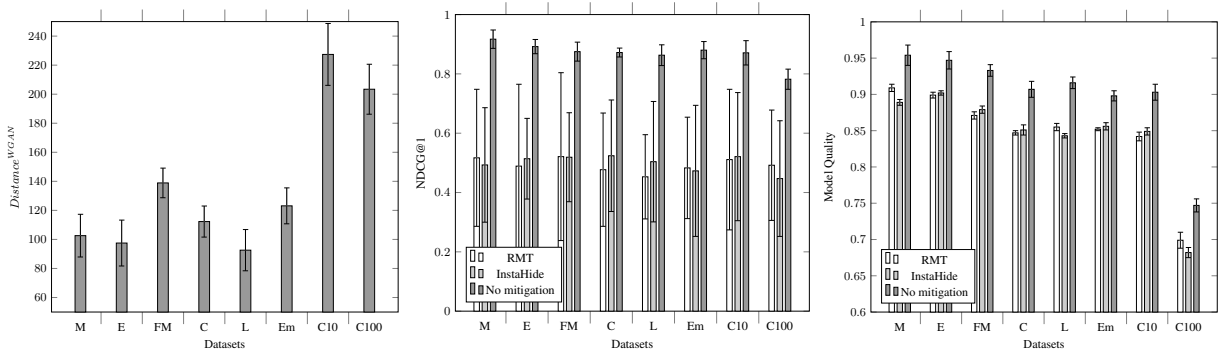

\subsection{Setup}
\textbf{Datasets.} We evaluate our method on eight well-known public datasets: MNIST, EMNIST, LFW, Emotion\footnote{the facial image data used by AffectNet \cite{EMOTION}}, CIFAR10, CIFAR100, Fashion-MNIST, and Clothing \cite{Clothing}, to make it easy to reproduce the results. We transform colored datasets, e.g., CIFAR datasets, to greyscale datasets and rescale all images to 36x36 pixels to unify the data formats. We also unify the number of classes with seven classes to align with the Emotion dataset which has only seven classes.

\textbf{Models.} For simplicity, we implement all the target models with the ResNet-18 architecture. We pick one of the eight datasets for each experiment to generate the target model and use the remaining as landmark datasets. We use the canonical WGAN \cite{Arjovsky17.1} for the GAN-based method.

\textbf{Training.} We use the 8:2 training-testing random split for each dataset and repeat it ten times for each specific experiment. We train the target models with the SGD optimizer with learning rate $10^{-2}$, batch size $100$, momentum $0.9$, and learning rate decay $10^{-4}$. To train the landmark WGAN, we randomly pick up $100$ images from every class and use the Adam optimizer with the learning rate $0.003$, batch size $20$, $\beta_1 = 0.5$ and $\beta_2 = 0.999$. To train the distracted WGAN, we set the target model weight $\lambda = 500$. Larger values will allow the target model to influence the distracted images more. $\lambda=500$ is determined experimentally to balance the target model's effect on the generator learning. We use the SGD optimizer with a learning rate of 0.02, batch size of 20, and momentum of 0.9 for training the distracted WGAN.

\subsection{Evaluation Metrics}
\label{ND}

\textbf{Dataset similarity.} As mentioned, we will use the OTDD dataset distance to represent the dataset similarity.

\textbf{Accuracy.} To evaluate the impact of auxiliary data on model-inversion attacks, we adopted the following method \cite{Zhang20}: we build up two reconstructed datasets with 700 images (100 images for every class) with and without auxiliary data, respectively, and then use the target model to classify the two datasets. The better the reconstruction quality, the higher accuracy the target model gives.

\textbf{NDCG score.} For each target model, we rank the landmark datasets by the corresponding distance measure in ascending order. The smaller distance, the more similar the landmark domain is to the target domain. To evaluate the ranking quality, we adopted the well-known measure: Normalized Discounted cumulative gain (NDCG) \cite{Croft10}.

Specifically, we first compute the pairwise distances between landmark datasets to derive the ground truth ranking per target domain.  
Let $R_i =  [d_{max} - d_{i,0},\dots, d_{max}-d_{i,n}]$ to be the ground truth ranking scores for the target domain $S_i$, where $d_{i,j}$ is the distance between $S_i$ and $S_j$ and $d_{max}$ is the maximum distance used to convert the distances to non-negative ranking scores (the larger the better). We define $R^{p}_i=[r_{max}-r_{i, 0},\dots,r_{max}-r_{i,n}]$, where $r_{i,j}$ is the inferred domain distance between $S_i$ and $S_j$ and $r_{max}$ is the largest inferred distance. Using the standard algorithm \cite{Croft10}, we can compute $NDCG_m$ for the top-m domains' ranking quality.

\subsection{Result Analysis}

\textbf{Effect of auxiliary data.}
In this set of experiments, we used the GAN-based attack to derive the domain similarity ranking. We choose the top-3 ranked datasets to show how they contribute to the model inversion attack. As shown in Figure \ref{F8}, the model inversion attack performs significantly better with the top-ranked datasets as the auxiliary data. Interestingly, the ranking is also consistent with their contributions to the model inversion attack for these top-ranked ones. Note that with or without the top-ranked auxiliary data, the attacking performance difference can be up to $\sim$30\%.

\textbf{Domain Inference Methods.}
We compare the GAN-based method with the model-inversion-based reconstruction method. For a clear presentation, we show only the results of NDCG\@1 to NDCG\@3. Specifically, NDCG\@1 looks at whether the top-1 result matches the ground truth, and NDCG\@m looks at the top-m result's ranking quality. Figure \ref{F2} shows that the GAN-based method performs consistently better than the reconstruction method for all datasets at all three levels. In Section \ref{GANb}, we mentioned using the differential domain similarity to eliminate the effect of GAN quality. Figure \ref{F9} shows that the differential similarity performs better consistently for every dataset than the distance between the original landmark dataset and the data generated by the distracted GAN (denoted as ``Alternative'' in the figure).

\textbf{Effect of WGAN quality.} We have observed that the MDI attack performs better on simpler datasets, e.g., MNIST and EMNIST than on the CIFAR datasets. A possible reason is that WGAN does not learn well on CIFAR datasets, also observed by previous studies \cite{Terjek19}. Figure \ref{fig:wgan-quality-and-mitigation} (left) clearly supports this observation. As the WGANs do not generate good-quality images, it's difficult to tell the distraction introduced by the target model or the GAN learning process. 

\textbf{Mitigation methods against MDI.}
In Section \ref{miti}, we have discussed several ways to protect models from the MDI attack. We are particularly interested in the low-cost model disguising methods, such as InstaHide \cite{Huang20} and RMT \cite{sharma21}. In this set of experiments, we experiment with these methods to see whether they can protect models from the domain inference attack and how much model quality we will need to trade-off. 

We use the RMT method to transform the training data with the setting of $block count = 4$ and $Noise level = 0$, i.e., each image is partitioned to four equal-size 16x16 blocks without noise addition. For InstaHide, we set $K=2$, i.e., mixing up the target image with one public image –- here, we randomly select a random image from any landmark datasets. Then, we use the new target models to repeat the experiments earlier. Figure \ref{fig:wgan-quality-and-mitigation} (mid) shows that both RMT and InstaHide significantly reduce the effectiveness of the MDI attack. Furthermore, the large error bars also show that the MDI attack results are volatile, indicating a strong level of protection. However, these data and model disguising methods will also slightly reduce the model quality. Figure \ref{fig:wgan-quality-and-mitigation} (right) shows the reduction is around 5\%. More sophisticated networks might close up this gap \cite{Huang20,sharma21}.

\section{Conclusion}
Most model-based attacks assume the domain knowledge is available to the adversary. In this paper, we study a critical problem: without explicitly knowing the model domain whether the attacker can effectively estimate the domain based on the model and any public or private datasets they can collect. We present a GAN-based model domain inference (MDI) method to infer similar domains of a black-box target model. Our approach aims to measure how the target model affects the training of a landmark domain's GAN model. The intuition is the more related/similar the target domain is to the landmark domain, the less the target model will disturb the GAN training. Our experimental results show that the proposed attack is highly effective in identifying similar domains: the auxiliary data from the top-ranked domains can significantly improve model-inversion attacks. We have empirically analyzed various factors affecting the effectiveness of the domain inference, including different architectures for estimating the domain similarity, the dataset similarity measures, and the effect of WGAN quality. We have also investigated the data and model disguising methods as a promising mitigation mechanism to protect the model from the MDI attack. 

\newpage
\bibliography{mda_papers.bib}

\begin{thebibliography}{32}
\providecommand{\natexlab}[1]{#1}

\bibitem[{Alvarez-Melis and Fusi(2020)}]{Alvarez-Melis20}
Alvarez-Melis, D.; and Fusi, N. 2020.
\newblock Geometric Dataset Distances via Optimal Transport.
\newblock arXiv:2002.02923.

\bibitem[{Arjovsky and Bottou(2017)}]{Arjovsky17.2}
Arjovsky, M.; and Bottou, L. 2017.
\newblock Towards principled methods for training generative adversarial
  networks.
\newblock \emph{arXiv preprint arXiv:1701.04862}.

\bibitem[{Arjovsky, Chintala, and Bottou(2017)}]{Arjovsky17.1}
Arjovsky, M.; Chintala, S.; and Bottou, L. 2017.
\newblock Wasserstein GAN.

\bibitem[{Carlini et~al.(2021)Carlini, Deng, Garg, Jha, Mahloujifar, Mahmoody,
  Thakurta, and Tram{\`e}r}]{Carlini21}
Carlini, N.; Deng, S.; Garg, S.; Jha, S.; Mahloujifar, S.; Mahmoody, M.;
  Thakurta, A.; and Tram{\`e}r, F. 2021.
\newblock Is private learning possible with instance encoding?
\newblock In \emph{2021 IEEE Symposium on Security and Privacy (SP)}, 410--427.
  IEEE.

\bibitem[{Chakraborty et~al.(2018)Chakraborty, Alam, Dey, Chattopadhyay, and
  Mukhopadhyay}]{chakraborty18}
Chakraborty, A.; Alam, M.; Dey, V.; Chattopadhyay, A.; and Mukhopadhyay, D.
  2018.
\newblock Adversarial Attacks and Defences: {A} Survey.
\newblock \emph{CoRR}, abs/1810.00069.

\bibitem[{Choquette-Choo et~al.(2021)Choquette-Choo, Tramer, Carlini, and
  Papernot}]{Choquette-Choo21}
Choquette-Choo, C.~A.; Tramer, F.; Carlini, N.; and Papernot, N. 2021.
\newblock Label-only membership inference attacks.
\newblock In \emph{International conference on machine learning}, 1964--1974.
  PMLR.

\bibitem[{Costan and Devadas(2016)}]{sgx-explained}
Costan, V.; and Devadas, S. 2016.
\newblock Intel SGX Explained.
\newblock \emph{IACR Cryptology ePrint Archive}, 2016: 86.

\bibitem[{Croft, Metzler, and Strohman(2010)}]{Croft10}
Croft, W.~B.; Metzler, D.; and Strohman, T. 2010.
\newblock \emph{Search engines: Information retrieval in practice}, volume 520.
\newblock Addison-Wesley Reading.

\bibitem[{Fredrikson, Jha, and Ristenpart(2015)}]{Fredrikson15}
Fredrikson, M.; Jha, S.; and Ristenpart, T. 2015.
\newblock Model Inversion Attacks That Exploit Confidence Information and Basic
  Countermeasures.
\newblock In \emph{Proceedings of the 22nd ACM SIGSAC Conference on Computer
  and Communications Security}, CCS '15, 1322--1333. New York, NY, USA:
  Association for Computing Machinery.
\newblock ISBN 9781450338325.

\bibitem[{Gilad-Bachrach et~al.(2016)Gilad-Bachrach, Dowlin, Laine, Lauter,
  Naehrig, and Wernsing}]{gilad16}
Gilad-Bachrach, R.; Dowlin, N.; Laine, K.; Lauter, K.; Naehrig, M.; and
  Wernsing, J. 2016.
\newblock CryptoNets: Applying Neural Networks to Encrypted Data with High
  Throughput and Accuracy.
\newblock In Balcan, M.~F.; and Weinberger, K.~Q., eds., \emph{Proceedings of
  The 33rd International Conference on Machine Learning}, volume~48 of
  \emph{Proceedings of Machine Learning Research}, 201--210.

\bibitem[{Grigorev(2020)}]{Clothing}
Grigorev, A. 2020.
\newblock Clothing dataset (full, high resolution).
\newblock
  \url{https://www.kaggle.com/datasets/agrigorev/clothing-dataset-full}.
\newblock Accessed: 2020-10-21.

\bibitem[{Heusel et~al.(2017)Heusel, Ramsauer, Unterthiner, Nessler, and
  Hochreiter}]{Heusel17}
Heusel, M.; Ramsauer, H.; Unterthiner, T.; Nessler, B.; and Hochreiter, S.
  2017.
\newblock GANs Trained by a Two Time-Scale Update Rule Converge to a Local Nash
  Equilibrium.

\bibitem[{Hidano et~al.(2017)Hidano, Murakami, Katsumata, Kiyomoto, and
  Hanaoka}]{Hidano17}
Hidano, S.; Murakami, T.; Katsumata, S.; Kiyomoto, S.; and Hanaoka, G. 2017.
\newblock Model inversion attacks for prediction systems: Without knowledge of
  non-sensitive attributes.
\newblock In \emph{2017 15th Annual Conference on Privacy, Security and Trust
  (PST)}, 115--11509. IEEE.

\bibitem[{Huang et~al.(2020)Huang, Song, Li, and Arora}]{Huang20}
Huang, Y.; Song, Z.; Li, K.; and Arora, S. 2020.
\newblock InstaHide: Instance-hiding Schemes for Private Distributed Learning.
\newblock \emph{CoRR}, abs/2010.02772.

\bibitem[{Hui et~al.(2021)Hui, Yang, Yuan, Burlina, Gong, and Cao}]{Hui21}
Hui, B.; Yang, Y.; Yuan, H.; Burlina, P.; Gong, N.~Z.; and Cao, Y. 2021.
\newblock Practical blind membership inference attack via differential
  comparisons.
\newblock \emph{arXiv preprint arXiv:2101.01341}.

\bibitem[{Khosravy et~al.(2021)Khosravy, Nakamura, Hirose, Nitta, and
  Babaguchi}]{Mahdi21}
Khosravy, M.; Nakamura, K.; Hirose, Y.; Nitta, N.; and Babaguchi, N. 2021.
\newblock Model inversion attack: analysis under gray-box scenario on deep
  learning based face recognition system.
\newblock \emph{KSII Transactions on Internet and Information Systems (TIIS)},
  15(3): 1100--1118.

\bibitem[{Li and Zhang(2020)}]{Li20}
Li, Z.; and Zhang, Y. 2020.
\newblock Label-leaks: Membership inference attack with label.
\newblock \emph{arXiv preprint arXiv:2007.15528}.

\bibitem[{Long et~al.(2020)Long, Wang, Bu, Bindschaedler, Wang, Tang, Gunter,
  and Chen}]{Long20}
Long, Y.; Wang, L.; Bu, D.; Bindschaedler, V.; Wang, X.; Tang, H.; Gunter,
  C.~A.; and Chen, K. 2020.
\newblock A pragmatic approach to membership inferences on machine learning
  models.
\newblock In \emph{2020 IEEE European Symposium on Security and Privacy
  (EuroS\&P)}, 521--534. IEEE.

\bibitem[{Mohassel and Zhang(2017)}]{Mohassel17}
Mohassel, P.; and Zhang, Y. 2017.
\newblock SecureML: A System for Scalable Privacy-Preserving Machine Learning.
\newblock In \emph{2017 IEEE Symposium on Security and Privacy (SP)}, 19--38.

\bibitem[{Mollahosseini, Hasani, and Mahoor(2017)}]{EMOTION}
Mollahosseini, A.; Hasani, B.; and Mahoor, M.~H. 2017.
\newblock Affectnet: A database for facial expression, valence, and arousal
  computing in the wild.
\newblock \emph{IEEE Transactions on Affective Computing}, 10(1): 18--31.

\bibitem[{Rahman et~al.(2018)Rahman, Rahman, Lagani{\`e}re, Mohammed, and
  Wang}]{Rahman18}
Rahman, M.~A.; Rahman, T.; Lagani{\`e}re, R.; Mohammed, N.; and Wang, Y. 2018.
\newblock Membership Inference Attack against Differentially Private Deep
  Learning Model.
\newblock \emph{Trans. Data Priv.}, 11(1): 61--79.

\bibitem[{Salem et~al.(2018)Salem, Zhang, Humbert, Berrang, Fritz, and
  Backes}]{Salem18}
Salem, A.; Zhang, Y.; Humbert, M.; Berrang, P.; Fritz, M.; and Backes, M. 2018.
\newblock Ml-leaks: Model and data independent membership inference attacks and
  defenses on machine learning models.
\newblock \emph{arXiv preprint arXiv:1806.01246}.

\bibitem[{Sharma, Alam, and Chen(2021)}]{sharma21}
Sharma, S.; Alam, A.~M.; and Chen, K. 2021.
\newblock Image Disguising for Protecting Data and Model Confidentiality in
  Outsourced Deep Learning.
\newblock In \emph{IEEE Conference on Cloud Computing}.

\bibitem[{Sharma and Chen(2019)}]{Sharma19SecureBoost}
Sharma, S.; and Chen, K. 2019.
\newblock Confidential boosting with random linear classifiers for outsourced
  user-generated data.
\newblock In \emph{European Symposium on Research in Computer Security},
  41--65. Springer.

\bibitem[{Shokri et~al.(2017)Shokri, Stronati, Song, and Shmatikov}]{Shokri17}
Shokri, R.; Stronati, M.; Song, C.; and Shmatikov, V. 2017.
\newblock Membership inference attacks against machine learning models.
\newblock In \emph{2017 IEEE symposium on security and privacy (SP)}, 3--18.
  IEEE.

\bibitem[{Song, Ristenpart, and Shmatikov(2017)}]{Song17}
Song, C.; Ristenpart, T.; and Shmatikov, V. 2017.
\newblock Machine Learning Models That Remember Too Much.
\newblock In \emph{Proceedings of the 2017 ACM SIGSAC Conference on Computer
  and Communications Security}, CCS '17, 587--601. New York, NY, USA:
  Association for Computing Machinery.
\newblock ISBN 9781450349468.

\bibitem[{Terj{\'e}k(2019)}]{Terjek19}
Terj{\'e}k, D. 2019.
\newblock Adversarial lipschitz regularization.
\newblock \emph{arXiv preprint arXiv:1907.05681}.

\bibitem[{Tramer and Boneh(2019)}]{Tramer19}
Tramer, F.; and Boneh, D. 2019.
\newblock Slalom: Fast, Verifiable and Private Execution of Neural Networks in
  Trusted Hardware.
\newblock In \emph{International Conference on Learning Representations}.

\bibitem[{Wang, Si, and Wu(2015)}]{Wang15}
Wang, Y.; Si, C.; and Wu, X. 2015.
\newblock Regression model fitting under differential privacy and model
  inversion attack.
\newblock In \emph{Twenty-fourth international joint conference on artificial
  intelligence}.

\bibitem[{Weng(2019)}]{Weng19}
Weng, L. 2019.
\newblock From gan to wgan.
\newblock \emph{arXiv preprint arXiv:1904.08994}.

\bibitem[{Yang, Chang, and Liang(2019)}]{Yang19}
Yang, Z.; Chang, E.-C.; and Liang, Z. 2019.
\newblock Adversarial neural network inversion via auxiliary knowledge
  alignment.
\newblock \emph{arXiv preprint arXiv:1902.08552}.

\bibitem[{Zhang et~al.(2020)Zhang, Jia, Pei, Wang, Li, and Song}]{Zhang20}
Zhang, Y.; Jia, R.; Pei, H.; Wang, W.; Li, B.; and Song, D. 2020.
\newblock The secret revealer: Generative model-inversion attacks against deep
  neural networks.
\newblock In \emph{Proceedings of the IEEE/CVF Conference on Computer Vision
  and Pattern Recognition}, 253--261.

\end{thebibliography}

\end{document}